\documentclass{article}
\usepackage[preprint]{neurips_2026}
\usepackage[utf8]{inputenc}
\usepackage[T1]{fontenc}
\usepackage{newunicodechar}
\newunicodechar{’}{\textquoteright}
\ifdefined\pdfminorversion
\fi

\usepackage{amsmath,amsfonts,bm}

\def\eqref#1{equation~\ref{#1}}

\def\1{\bm{1}}

\DeclareMathAlphabet{\mathsfit}{\encodingdefault}{\sfdefault}{m}{sl}
\SetMathAlphabet{\mathsfit}{bold}{\encodingdefault}{\sfdefault}{bx}{n}

\setcitestyle{authoryear,round,citesep={;},aysep={,},yysep={;}}

\usepackage{graphicx}
\usepackage{booktabs}
\usepackage[table]{xcolor}
\usepackage{tabularx}
\usepackage{multirow}
\usepackage{pifont}
\usepackage[breakable]{tcolorbox}
\usepackage{fvextra}
\usepackage[varqu]{zi4}
\usepackage{url}
\usepackage[pdfversion=1.7,colorlinks=true,linkcolor=blue,citecolor=blue,urlcolor=blue]{hyperref}
\usepackage{cleveref}

\crefname{table}{Tab.}{Tabs.}
\Crefname{table}{Table}{Tables}
\crefname{figure}{Fig.}{Figs.}
\Crefname{figure}{Figure}{Figures}
\crefname{section}{Sec.}{Secs.}
\Crefname{section}{Section}{Sections}

\title{Do LLMs Act on What They Know? From Partner Representations to Cooperative Actions}
\author{%
  Yuhwan Jeong\thanks{Equal contribution.} \quad Jinnyeong Yang\footnotemark[1] \quad Kuk-Jin Yoon \\
  KAIST, Visual Intelligence Lab \\
  \texttt{\{jeongyh98,jinnyeong6118,kjyoon\}@kaist.ac.kr}
}
\hypersetup{pdfauthor={Yuhwan Jeong, Jinnyeong Yang, Kuk-Jin Yoon},pdftitle={Do LLMs Act on What They Know? From Partner Representations to Cooperative Actions}}

\begin{document}

\maketitle

\begin{abstract}
Cooperation with unfamiliar partners requires adapting to communication conventions that are not known in advance.
We study this problem in a controlled Hanabi-derived environment with scripted hint generation, LLM-controlled receiving decisions, and frozen model weights.
Across eight LLMs, linear probes recover intent conventions substantially more accurately than target conventions, yet receiving choices do not consistently agree with the sender's convention.
We compare probe-predicted and ground-truth conventions presented either as general rules or as externally computed action recommendations.
Rule statements yield modest and model-dependent changes in cooperation, whereas action translation produces larger gains on average.
In a Qwen3-8B case study, matched-state statement reversals reveal much greater sensitivity to action recommendations than to rule statements.
Activation transfers from oracle-action and non-oracle hint-restatement donors improve intent accuracy on both action classes, but the tested alternatives do not reliably reproduce these benefits.
Together, these results distinguish convention decodability, sensitivity to convention information, and cooperative performance, and highlight limitations in turning available partner information into receiving decisions.
\end{abstract}
\section{Introduction}
\label{sec:intro}

Large language models (LLMs) are increasingly deployed as decision makers that carry out tasks with humans or other agents.
In these settings, success depends not only on solving a task individually but also on coordinating choices with a partner.
A model must account for what its partner intends, how the partner acts, and how its own actions will be interpreted.
Even LLMs with strong task knowledge and reasoning abilities can struggle with these demands~\citep{agashe2025llmcoordination, ramesh2026sparks, liang2025llmhanabi, yao2025spinbench}.
Understanding LLM cooperation therefore requires examining how information about a partner informs the model's decisions.

This requirement becomes particularly important when cooperating with an unfamiliar partner.
Partners may share a goal while using different strategies or communication conventions, making a previously effective response inappropriate in a new interaction~\citep{bard2020hanabi, carroll2019utility}.
An agent must use observed behavior to infer relevant partner characteristics and adjust its subsequent choices.
Zero-shot coordination and ad hoc teamwork provide important settings for studying cooperation with partners that were not encountered during joint training~\citep{hu2020otherplay, dizdarevic2025adhoc}.
Within this broader problem, we focus on how LLMs use interaction history to adapt their behavior without partner-specific updates to the base model's weights.

Prior work incorporates explicit reasoning about partners into LLM planning~\citep{li2023theory, zhang2024coela}.
ProAgent estimates teammate intentions and updates its beliefs from observed behavior~\citep{zhang2024proagent}, while Hypothetical Minds generates, evaluates, and revises natural-language hypotheses about partner strategies~\citep{cross2025hypothetical}.
However, accurate partner prediction need not translate into behavioral adaptation~\citep{riemer2025theory}.
In communication-based cooperation, this distinction matters because interpreting a signal may require applying an estimate of the partner's convention.
For LLM agents, behavioral adaptation therefore depends not only on forming such an estimate, but also on whether that information influences the final decision.
We ask whether partner conventions are decodable from the model's hidden representations and whether making that information available changes how the model interprets the current hint.

\begin{figure}[t]
  \centering
  \includegraphics[width=0.99\linewidth]{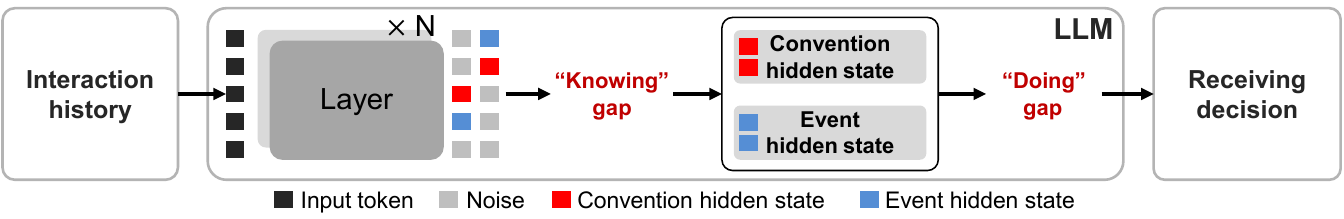}
  \vspace{-10pt}
  \caption{\textbf{Conceptual overview from interaction history to receiving decision.}
\emph{Knowing} measures how accurately the sender's convention can be decoded from the receiver's hidden states, while \emph{Doing} measures whether the receiver's target and intent choices agree with the sender's convention.
The internal components are schematic rather than identified model subspaces.}
  \vspace{-10pt}
  \label{fig:fig1_pipeline}
\end{figure}

To study these questions, we use a controlled environment derived from Hanabi in which agents have different communication conventions~\citep{bard2020hanabi}.
The same hint can refer to different cards or request different decisions depending on the sender's convention.
Each agent retains its own sending convention while receiving hints from a partner whose convention is not explicitly disclosed.
We collect probing data from scripted interactions and evaluate interventions in games where language models make receiving decisions while hint generation remains scripted.
This design focuses the analysis on convention-dependent interpretation and action selection while controlling the sending policy.



To distinguish convention inference from its application, we probe the receiver's hidden representations for the partner's assigned convention~\citep{hewitt2019designing,belinkov2022probing}.
We then evaluate textual and activation-based interventions to test whether decodable convention information can influence action selection, while keeping the base LLM frozen throughout~\citep{zhang2024activation,rimsky2024steering}.
We separately evaluate \emph{knowing}, whether the sender's convention can be decoded from the receiver's representations, and \emph{doing}, whether the receiver's choices follow that convention.
We report game score separately as the resulting cooperative outcome.

Our experiments show that partner-convention information is more readily decoded on the intent axis than on the target axis, yet receiving decisions do not consistently follow the partner's convention.
Providing convention information as a rule has limited effects on cooperation, whereas translating it into the current decision yields larger gains.
Matched-state analyses show much stronger responses to action recommendations than to rule statements.
Activation-based analyses further show that some state-specific donor transfers improve both action classes, whereas simpler transformations do not preserve these gains.
Our contributions are threefold:
\begin{itemize}
\item We develop a controlled cooperation setting that separates sender-specific convention inference, interpretation of the current hint, and receiving decisions.
\item We show that convention information presented as a rule has limited effects on cooperation, whereas translating it into action recommendations yields larger gains.
\item We use matched-state and activation-based analyses to distinguish decodable partner information from its behavioral use and from changes in default action preference.
\end{itemize}

\section{Related Work}
\label{sec:related}

\paragraph{Partner-aware coordination.}
Successful cooperation requires agents to coordinate with partners whose strategies, conventions, or preferences may differ from their own.
Prior work has studied robust coordination through symmetry-based training~\citep{hu2020otherplay}, cross-environment and population objectives~\citep{jha2025crossenvironment, hui2026efficient}, and general formulations for unseen teammates~\citep{dizdarevic2025adhoc}.
A related line grounds cooperation in human behavior, from imitation-based methods~\citep{carroll2019utility, jacob2022pikl, bakhtin2022cicero} to influence-based team steering~\citep{sheng2026influence}.
More recently, LLM agents have been evaluated in cooperative games~\citep{agashe2025llmcoordination, liang2025llmhanabi, ramesh2026sparks} and equipped with explicit partner models~\citep{zhang2024proagent, cross2025hypothetical, li2023theory, zhang2024coela, goel2026bayesian}.
Related work studies in-context co-player inference in sequence-model agents~\citep{weis2026multiagent} and belief tracking and partner generalization in multi-agent environments~\citep{ruhdorfer2026yokai}.
Other work questions what behavioral evaluations alone establish~\citep{riemer2025theory, chandak2025answer, singh2025pitfalls}.
Behavioral performance alone does not reveal whether partner information is absent or simply unused.
Our interventions instead ask whether feeding the readout back changes decisions in line with the partner's convention rather than merely shifting default action tendencies.

\paragraph{Probing internal representations.}
Probing studies have recovered latent world-state information from game-playing models~\citep{li2023emergent, nanda2023emergent, karvonen2024emergent, kamel2025emergent} and cooperative policies~\citep{tessera2026probing}, while belief representations have been shown to support intervention~\citep{sarfati2026shape}.
Related work also identifies internal partner models~\citep{monwilliams2025partner}, belief-tracking mechanisms~\citep{prakash2026lookbacks}, and discrepancies between internal or intermediate information and final behavior~\citep{lekeas2026suppresses, young2026know, boppana2026theater, li2026performative, arghal2026behavioural}.
Because probing can suffer from confounds, unselective readouts, and statistical instability~\citep{hewitt2019designing, belinkov2022probing, nordby2026linear, meloux2025mechanistic}, we pair each readout with a behavioral intervention on the same decisions, comparing whether partner information is readable and whether it influences behavior.

\paragraph{Activation patching and steering.}
Activation patching tests causal structure by transplanting internal states between runs~\citep{zhang2024activation}, with recent work extending it to circuit-level analysis~\citep{haklay2025position, wu2026demystifying}. Steering instead adds learned or synthesized directions at inference time~\citep{rimsky2024steering}, including applications to game strategies and theory-of-mind behavior~\citep{sun2026persona, li2026costom}, though its robustness remains an active concern~\citep{tan2024analysing, im2025unified, braun2026beyond, nguyen2026generalization}.

\section{Task and Experimental Setup}
\label{sec:env}

\subsection{Convention-Based Cooperation Task}
\label{sec:instrument}

We study how LLMs use interaction history to interpret a partner's communication and adapt their choices.
Our environment is derived from Hanabi, a cooperative card game in which each player observes the partner's hand but not their own~\citep{bard2020hanabi}.
Players can play a card, discard a card, or spend a hint token to identify all cards of a particular color or rank in the partner's hand.
The game rules determine which cards a hint marks, while a communication convention specifies the intended response.
We use three colors with five ranks per color, so completing all color stacks yields a maximum shared score of 15.
Playing a card that is not the next rank in its color stack costs one shared life, and the episode terminates when no lives remain.
The environment is implemented using the Hanabi Learning Environment, with game parameters listed in Appendix~\ref{app:constants}.


Each player $P \in \{A, B\}$ receives a sending convention
$\theta_P=(\theta_P^{\mathrm{target}},\theta_P^{\mathrm{intent}})$
that remains fixed within an episode.
The target convention specifies whether a hint refers to the leftmost or rightmost marked card.
The intent convention specifies whether color hints mean \textsc{Play} and rank hints mean \textsc{Discard}, or vice versa.
On each received hint, the receiver then makes two decisions: a target decision, which selects the marked card intended by the sender, and an intent decision, which selects \textsc{Play} or \textsc{Discard}.
We evaluate the two decisions separately.
Each player knows its own convention but is not explicitly told the partner's convention.

\begin{figure}[t]
  \centering
  \includegraphics[width=0.99\linewidth]{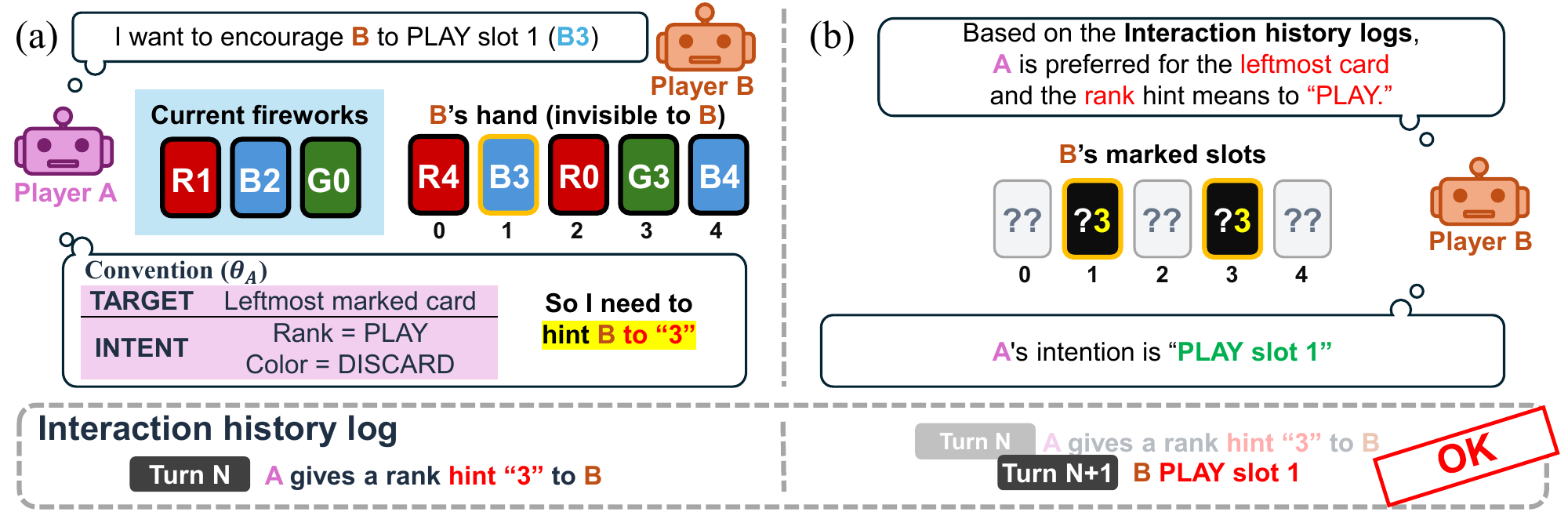}
  \vspace{-10pt}
  \caption{Sender-specific communication conventions.
  (a) Player $A$ wants $B$ to play slot 1 and, under its target convention (leftmost marked card) and intent convention (rank means \textsc{Play}), gives the rank hint 3, which marks slots 1 and 3.
  (b) Player $B$ sees only the interaction log, infers $A$'s convention from earlier hints and reactions, and interprets the hint as \textsc{Play} on slot 1.}
  \label{fig:play}
  \vspace{-10pt}
\end{figure}

When $A$ sends a hint to $B$, its meaning is determined by $\theta_A$, which $B$ must infer from interaction history.
The roles reverse when $B$ sends a hint to $A$.
Interpreting a received hint therefore requires combining the sender's convention with the marked slots and hint type, as~\cref{fig:play} (a) illustrates for a single hint.
Both plays and discards reveal cards that can help the receiver assess earlier hints, although individual events may remain ambiguous.
In particular, a successful play does not by itself establish that the receiver selected the card intended by the sender.

\subsection{Interaction Settings}
\label{sec:interaction}


We use scripted interactions between bots and evaluate receiving interventions in controlled games.
Each scripted bot sends hints according to its assigned convention and interprets received hints using that same convention. Because the two bots' conventions are drawn independently, the resulting histories contain both reactions that match the sender's intended meaning and reactions that do not.
In the controlled games, hint generation is scripted for both players, while, as illustrated in \cref{fig:play}(b), LLMs make two receiving decisions at designated turns, selecting the target slot and choosing \textsc{Play} or \textsc{Discard}. The surrounding policy controls turn selection and actions outside these decisions. We keep this policy and each player's assigned convention fixed across receiving interventions. This setting isolates receiving decisions rather than evaluating fully autonomous game play.

\subsection{Model Interface}
\label{sec:prompts}



The base language model’s weights remain fixed throughout evaluation. Each prompt contains the task rules, the acting player’s own convention, the interaction history, and the game state visible to that player.
A NOTES block is present in every condition and contains additional information only when provided by the receiving condition.
Target questions offer the distinct leftmost and rightmost marked slots, while intent questions offer \textsc{Play} and \textsc{Discard} for a specified slot. The two options are labeled A and B, and we select the label with the higher next-token logit at the answer position. In game evaluations, the option order is randomized across decisions. Data partitions and prompt templates are provided in Appendices~\ref{app:constants} and~\ref{app:format}.

\section{Probing and Interventions}
\label{sec:method}

Our approach combines linear probes of the partner's convention with textual and activation-based interventions on the receiver's decisions.
Oracle and action-translation conditions provide diagnostic comparisons that vary information accuracy and the processing performed outside the model.

\subsection{Convention Probes}
\label{sec:positions}
\label{sec:gates}

For an interaction from sender $A$ to receiver $B$, we predict $A$'s assigned convention from hidden representations of $B$'s observation.
At token position $t$, we concatenate the embedding output and the hidden states from all transformer layers, following work that combines layers to read information distributed across depth~\citep{devlin2019bert,tenney2019bert}, and fit a separate linear logistic probe for each convention axis:
\begin{equation}
    x_t = [h_t^{(0)};\ldots;h_t^{(L)}],
    \qquad
    p_a(\theta_A^a = 1 \mid x_t)
    = \sigma(w_a^\top x_t + b_a),
\end{equation}
where $a \in \{\mathrm{target}, \mathrm{intent}\}$ and $h_t^{(0)}$ is the embedding output.
Labels identify the sender's convention, independently of the current marked slots and option order.
Probes are trained separately for each language model on the scripted interactions of~\cref{sec:interaction}, pooling examples from both players while preserving each receiver's observation limits.


Our primary readout, \emph{last-token}, uses the final token of the prompt body, before the action question and options are appended, so it reads the observation provided for action selection without a question eliciting the partner's convention.
We use it for both axes and for the readout-guided interventions below.
As an auxiliary analysis, the \emph{event-aggregated} readout combines predictions from the final tokens of the history lines that record hints received from the partner and actions taken by the receiver.
The selection rule and the training settings are in Appendix~\ref{app:format}.

\subsection{Convention Information and Action Selection}
\label{sec:stage}




We vary two factors: the source of convention information and whether it is translated into the current target and intent decisions.
Prediction-based conditions use the last-token probe's estimate of the sender's convention, whereas oracle conditions use the sender's assigned convention.

In \texttt{loop}, the probe-predicted target and intent conventions are fed back into NOTES as general rules.
The model must then apply these rules to the current hint to make a target decision and an intent decision.
The oracle counterpart, \texttt{told}, uses the sender's assigned convention in the same rule-statement form.
The suffix \texttt{verb} denotes action translation.
In \texttt{loop + verb}, external code applies the probe-predicted convention to the current hint and provides the resulting target slot and intent decision in NOTES.
Likewise, \texttt{told + verb} applies the sender's assigned convention to the current hint.
For example, if a rank hint marks slots 1 and 3 and the supplied convention specifies the leftmost marked card and rank as \textsc{Play}, \texttt{loop} states those two rules, whereas \texttt{loop + verb} recommends slot 1 and \textsc{Play}.
The LLM still makes the final choice in all conditions.

The \texttt{raw} condition receives no partner-convention information.
Thus, the oracle conditions vary convention accuracy, while the \texttt{+ verb} conditions reduce the convention-to-decision mapping left to the model.
The exact wording is given in Appendix~\ref{app:format}.

\subsection{Activation Interventions}
\label{sec:activation_interventions}



We compare donor activation transfers with transformations of the receiver's representations.
Donor runs either provide oracle action information or restate the current hint alongside the predicted convention.
We transfer attention-head outputs at the answer position and separately test donor attention weights with recipient value vectors~\citep{zhang2024activation}.

We also evaluate several transformations of the receiver's activations, including head-output scaling, convention-contrast injection, fitted linear maps, and averaged restatement directions~\citep{rimsky2024steering}.
Additional diagnostics test attention to the current hint, query replacement, and a minimal-context self-donor.
Matched comparisons hold the state, queried slot, and answer options fixed.
The base LLM remains frozen except in the learning diagnostic of~\cref{sec:lora}.
Intervention details and implementation checks are in Appendix~\ref{app:activation_interventions}.

\section{Results}
\label{sec:results}

\subsection{Models and Evaluation}
\label{sec:setting}


We evaluate eight open-weight LLMs: Qwen3-8B and Qwen3-32B~\citep{yang2025qwen3}, R1-Distill-Qwen-7B~\citep{deepseek2025r1}, Llama-3.1-8B~\citep{grattafiori2024llama3}, Tulu-3-8B~\citep{lambert2025tulu3}, Hermes-3-8B~\citep{teknium2024hermes3}, Gemma-3-12B~\citep{gemma2025gemma3}, and Phi-4~\citep{abdin2024phi4}.
We abbreviate R1-Distill-Qwen-7B as R1-7B in figures and tables.
A separate convention probe is trained for each model.
We develop the experimental setup on Qwen3-8B and evaluate the probe readouts and prompt-level conditions on all eight LLMs.
The matched-state and activation analyses focus on Qwen3-8B.
We report \emph{knowing}, \emph{doing}, and game score separately.
Game-level comparisons use matched seeds and convention configurations, while state-level analyses compare interventions on identical recorded receiving states.

\begin{table}[t]
\centering\small
\caption{Knowing across eight LLMs, the probe accuracy for the sender's target and intent conventions in scripted interactions and controlled games.
Last and Event are the last-token and event-aggregated readouts, Turn-avg. averages over turns, Final uses the last turn of each game.}
\label{tab:knowing}
\vspace{2pt}
\setlength{\tabcolsep}{4pt}
\begin{tabular}{ll cccc cccc}
\toprule
& & \multicolumn{4}{c}{Target} & \multicolumn{4}{c}{Intent} \\
\cmidrule(lr){3-6}\cmidrule(lr){7-10}
& & \multicolumn{2}{c}{Scripted} & \multicolumn{2}{c}{Games} & \multicolumn{2}{c}{Scripted} & \multicolumn{2}{c}{Games} \\
\cmidrule(lr){3-4}\cmidrule(lr){5-6}\cmidrule(lr){7-8}\cmidrule(lr){9-10}
Model & Readout & Turn-avg. & Final & Turn-avg. & Final & Turn-avg. & Final & Turn-avg. & Final \\
\noalign{\global\aboverulesep=0.4ex \global\belowrulesep=0pt}
\midrule
\rowcolor{gray!12}
\multicolumn{10}{l}{Qwen family} \\
\noalign{\global\aboverulesep=0pt \global\belowrulesep=0.65ex}
\midrule
\multirow{2}{*}{Qwen3-8B} & Last & 0.638 & 0.677 & 0.569 & 0.596 & 0.921 & 0.965 & 0.814 & 0.873 \\
 & Event & 0.659 & 0.747 & 0.607 & 0.658 & 0.964 & 0.972 & 0.869 & 0.883 \\
\multirow{2}{*}{Qwen3-32B} & Last & 0.620 & 0.691 & 0.564 & 0.623 & 0.916 & 0.927 & 0.813 & 0.840 \\
 & Event & 0.655 & 0.736 & 0.585 & 0.637 & 0.970 & 0.990 & 0.892 & 0.925 \\
\multirow{2}{*}{R1-7B} & Last & 0.612 & 0.604 & 0.499 & 0.506 & 0.867 & 0.885 & 0.782 & 0.796 \\
 & Event & 0.637 & 0.726 & 0.539 & 0.588 & 0.963 & 0.976 & 0.883 & 0.894 \\
\noalign{\global\aboverulesep=0.4ex \global\belowrulesep=0pt}
\midrule
\rowcolor{gray!12}
\multicolumn{10}{l}{Llama family} \\
\noalign{\global\aboverulesep=0pt \global\belowrulesep=0.65ex}
\midrule
\multirow{2}{*}{Llama-3.1-8B} & Last & 0.632 & 0.653 & 0.504 & 0.498 & 0.911 & 0.955 & 0.795 & 0.819 \\
 & Event & 0.658 & 0.708 & 0.534 & 0.577 & 0.961 & 0.979 & 0.830 & 0.831 \\
\multirow{2}{*}{Tulu-3-8B} & Last & 0.627 & 0.656 & 0.500 & 0.521 & 0.911 & 0.944 & 0.794 & 0.823 \\
 & Event & 0.638 & 0.708 & 0.549 & 0.594 & 0.970 & 0.983 & 0.838 & 0.842 \\
\multirow{2}{*}{Hermes-3-8B} & Last & 0.616 & 0.625 & 0.499 & 0.535 & 0.909 & 0.948 & 0.815 & 0.835 \\
 & Event & 0.660 & 0.705 & 0.538 & 0.592 & 0.964 & 0.972 & 0.879 & 0.892 \\
\noalign{\global\aboverulesep=0.4ex \global\belowrulesep=0pt}
\midrule
\rowcolor{gray!12}
\multicolumn{10}{l}{Others} \\
\noalign{\global\aboverulesep=0pt \global\belowrulesep=0.65ex}
\midrule
\multirow{2}{*}{Gemma-3-12B} & Last & 0.649 & 0.688 & 0.510 & 0.523 & 0.909 & 0.962 & 0.822 & 0.867 \\
 & Event & 0.666 & 0.760 & 0.523 & 0.548 & 0.940 & 0.948 & 0.844 & 0.850 \\
\multirow{2}{*}{Phi-4} & Last & 0.595 & 0.646 & 0.506 & 0.531 & 0.907 & 0.958 & 0.838 & 0.875 \\
 & Event & 0.635 & 0.705 & 0.563 & 0.590 & 0.959 & 0.969 & 0.869 & 0.879 \\
\bottomrule
\end{tabular}
\vspace{-10pt}
\end{table}

\begin{figure}[!t]
\centering
\includegraphics[width=0.99\linewidth]{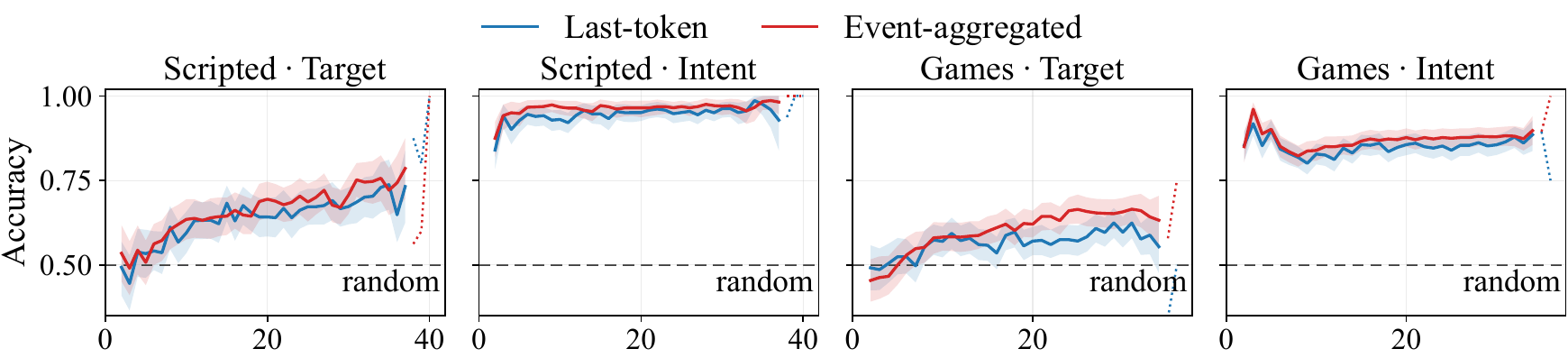}
\vspace{-10pt}
\caption{Probe accuracy by turn for Qwen3-8B.
Bands are 95\% confidence intervals over games, and a curve turns dotted where few games remain.}
\label{fig:turncurve}
\vspace{-10pt}
\end{figure}

\subsection{Knowing}
\label{sec:results-knowing}

We first measure knowing, the accuracy with which a probe recovers the sender's assigned convention from the receiver's hidden states, for every LLM on scripted interactions and in controlled games, and report it in~\cref{tab:knowing}.
The last-token readout is our primary measure and the event-aggregated readout an auxiliary comparison, and for each axis of $\theta$ they predict
\begin{equation}
\hat{\theta}_{\mathrm{last}} = \operatorname{sign} f_{\mathrm{last}}(h_{\mathrm{last}}),
\qquad
\hat{\theta}_{\mathrm{event}} = \operatorname{sign} \sum_{i \in E_t} f_{\mathrm{event}}(h_i),
\label{eq:readouts}
\end{equation}
where $h_{\mathrm{last}}$ is the hidden state at the final token of the prompt body, $E_t$ the history lines up to turn $t$ that record hints the partner gave and actions the receiver took, $h_i$ the hidden state at the final token of line $i$, and $f_{\mathrm{last}}$ and $f_{\mathrm{event}}$ probe logits trained at those positions.
Across LLMs, intent is more accurately decoded than target in both settings.
For Qwen3-8B in controlled games, turn-averaged last-token accuracy is 0.569 for target and 0.814 for intent, compared with 0.607 and 0.869 for the event-aggregated readout.
Because the readouts differ in token position, event selection, and aggregation, their accuracy difference does not isolate the contribution of external aggregation.

We also track the same accuracy over turns for Qwen3-8B in~\cref{fig:turncurve}.
Target accuracy generally rises over turns, whereas intent accuracy is high early in the interaction.
Later turns contain different board states and fewer surviving games, so each point is the accuracy over the games that reach that turn, not the accuracy of the same games measured at an earlier and a later turn.
The per-turn counts and the availability of event-aggregated predictions are in Appendix~\ref{app:turncurve}.
These results show that convention information is accessible to the trained probes, but do not establish that the model uses it when selecting actions~\citep{hewitt2019designing,belinkov2022probing}.

\subsection{Doing}
\label{sec:results-doing}

We next measure doing, the rate at which the receiver's target and intent decisions agree with the sender's convention.
We report it together with game score under the five receiving conditions of Qwen3-8B in~\cref{tab:doing}.
Without convention information (\texttt{raw}), target and intent doing are 0.515 and 0.591, and the predicted convention as a rule statement (\texttt{loop}) moves them only to 0.531 and 0.612, with a score increase of $+0.17$.


The two axes respond differently to information accuracy and action translation.
Applying the same prediction to the current hint (\texttt{loop + verb}) raises intent doing from 0.612 to 0.760 while target doing remains at 0.510, with a score increase of $+0.34^{*}$ over \texttt{loop}.
Supplying the true convention as a rule statement (\texttt{told}) instead raises target doing to 0.821 while intent doing remains at 0.599, with a score increase of $+0.36^{*}$.
Thus, target choices improve with a corrected convention, whereas intent choices respond to its translation into the current target and intent decisions.
Combining both (\texttt{told + verb}) yields the highest doing on both axes, at 0.962 for target and 0.867 for intent, and a game score of 7.78, $+2.15^{*}$ over \texttt{raw}.
Code that follows the sender's convention exactly scores 8.45, so \texttt{told + verb} remains $0.67^{*}$ below the ceiling of this environment and sending policy.
These results show that both information accuracy and action translation affect receiving behavior and cooperation.


\begin{table}[t]
\centering\small
\caption{Doing and game score for Qwen3-8B under the five receiving conditions.
A rule statement leaves its application to the LLM, whereas action translation provides the target slot and intent decision.
Ceiling replaces the LLM by code that follows the sender's convention exactly.}
\label{tab:doing}
\vspace{2pt}
\begin{tabular}{lllrrr}
\toprule
Condition & Convention source & Provided information & Doing (target) & Doing (intent) & Score \\
\midrule
\texttt{raw} & --- & --- & 0.515 & 0.591 & 5.63 \\
\addlinespace
\texttt{loop} & \multirow{2}{*}{Probe prediction}
 & Rule statement & 0.531 & 0.612 & 5.80 \\
\texttt{loop + verb} &
 & Action translation & 0.510 & 0.760 & 6.14 \\
\addlinespace
\texttt{told} & \multirow{2}{*}{Ground truth}
 & Rule statement & 0.821 & 0.599 & 6.16 \\
\texttt{told + verb} &
 & Action translation & 0.962 & 0.867 & 7.78 \\
\addlinespace
Ceiling & Ground truth & --- & 1.000 & 1.000 & 8.45 \\
\bottomrule
\end{tabular}
\vspace{-10pt}
\end{table}

\subsection{Self-Play and Cross-Play}
\label{sec:crossplay}

\begin{figure}[t]
\centering
\includegraphics[width=0.99\linewidth]{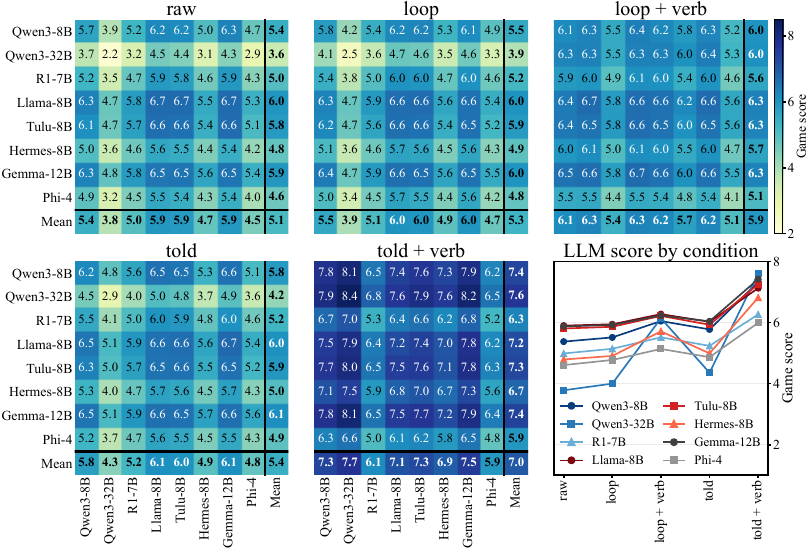}
\vspace{-10pt}
\caption{Self-play and cross-play game scores across LLMs and conditions.
Rows give the LLM playing as $A$, which moves first, and columns the LLM playing as $B$.
Mean entries average the eight cells of a row or column.
The lower-right panel gives each LLM's mean score under each condition.}
\vspace{-15pt}
\label{fig:crossplay}
\end{figure}

We then run every LLM with an unassisted receiver and report doing and game score in~\cref{fig:deployment}; the exact values are in Appendix~\ref{app:model-conditions}.
We further pair the eight LLMs in self-play and cross-play under the same scripted sending policy and report the scores in~\cref{fig:crossplay}.
Self-play pairs two instances of the same LLM, while cross-play pairs different LLMs.
Each ordered pairing is evaluated on 240 games across four convention configurations.



Across the 56 ordered cross-play pairings, the mean score is 5.1 for \texttt{raw}, 5.3 for \texttt{loop}, and 5.4 for \texttt{told}.
Providing rule statements therefore produces relatively small aggregate changes, even when the conventions are correct.
In contrast, \texttt{told + verb} reaches a cross-play mean of 7.0.
Larger gains from action translation also appear in several self-play comparisons.
For example, Qwen3-32B scores 2.2 under \texttt{raw}, 2.9 under \texttt{told}, and 8.4 under \texttt{told + verb}.
Thus, the effect of translating convention information into the current decision extends beyond Qwen3-8B.
Doing and game score under every condition for each LLM are reported in Appendix~\ref{app:model-conditions}.

A similar aggregate pattern appears without ground truth.
Translating the predicted convention with \texttt{loop + verb} raises the cross-play mean from 5.3 to 5.9.
For Qwen3-32B paired with Qwen3-8B, the score rises from 4.2 and 4.1 under \texttt{loop} to 6.3 with either model as $A$.
These results show that how convention information is presented to the receiver matters across LLM pairings.
Game score alone, however, does not reveal whether the receiver became more sensitive to the convention or mainly shifted its action preference.
We examine this distinction with matched-state analyses next.


\begin{figure}[t]
\centering
\includegraphics[width=0.99\linewidth]{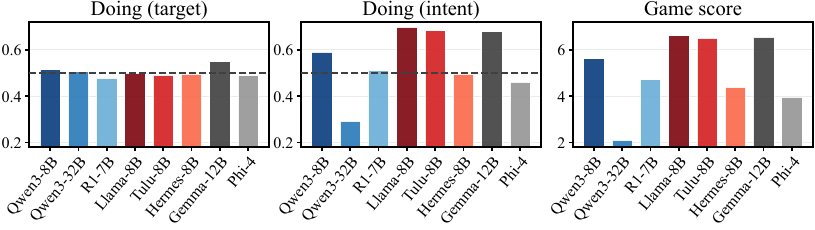}
\vspace{-10pt}
\caption{Doing and game score across LLMs with an unassisted receiver (\texttt{raw}), higher is better.
The dashed line at 0.5 marks random choice.
Bar colors mark the model family.}
\vspace{-10pt}
\label{fig:deployment}
\end{figure}

\subsection{Responses to Convention Information}
\label{sec:convention_response}

To separate the two, we reverse the notes statement on 800 recorded receiving states of Qwen3-8B, 348 with a color hint and 452 with a rank hint, while holding the history, the selected slot, the question, and the answer options fixed, once for the rule statement and once for a direct action recommendation, and report the responses in~\cref{tab:statement_response}.
We focus on the intent axis, where a true rule statement did not change behavior in~\cref{tab:doing}, and ask whether the receiver responds to the content of the statement at all.

\begin{table}[t]
\centering
\small
\caption{Sensitivity to statement reversals on 800 matched Qwen3-8B intent decisions.
Mean aligned $\Delta$ measures the logit shift in the implied direction.
Aligned fraction, choice-flip rate, and zero-shift rate report directional shifts, changed choices, and exactly zero shifts, respectively.}
\label{tab:statement_response}
\vspace{2pt}
\setlength{\tabcolsep}{5pt}
\begin{tabular}{lcccc}
\toprule
Statement & Mean aligned $\Delta$ & Aligned fraction & Choice-flip rate & Zero-shift rate \\
\midrule
Rule statement (color hint) & 0.157 & 0.91 & 0.135 & 0.23 \\
Rule statement (rank hint) & $-0.005$ & 0.47 & 0.080 & 0.34 \\
Rule statement (all) & 0.066 & 0.68 & 0.104 & 0.29 \\
Action recommendation & 1.378 & 1.00 & 0.586 & 0.00 \\
\bottomrule
\end{tabular}
\vspace{-15pt}
\end{table}


Reversing the action recommendation moves the answer logit by 1.378 on average, always in the implied direction, and changes the selected action in 59\% of states.
Reversing the rule statement moves it by only 0.066, changes the selection in 10\% of states, and leaves the logits unchanged in 29\%.
Its remaining effect is concentrated in color hints: 91\% of moving states shift in the implied direction, versus 47\% for rank hints.
The receiver therefore follows an explicit action recommendation but applies a stated rule weakly and mainly for one hint type.
Aggregate accuracy can hide this distinction: scaling one attention head raises intent accuracy from 0.584 to 0.605 only by choosing \textsc{Play} more often, helping \textsc{Play} states while harming \textsc{Discard} states.
We therefore report intervention effects separately for states requiring \textsc{Play} and \textsc{Discard}; the precision check and scaling results are in Appendix~\ref{app:convention_response}.

\subsection{Effects of Activation Interventions}
\label{sec:activation_effects}

Since a stated rule has little effect on the answer, we ask whether activation interventions can better support the convention-to-decision mapping.
We try this in two ways on Qwen3-8B's receiving activations, transferring attention-head outputs from donor runs that provide more explicit decision information, and applying fixed transformations without a donor run, and report the results in~\cref{tab:activation_effects}.

\begin{table}[t]
\centering
\small
\caption{Activation-intervention effects on intent accuracy, split by the true intent decision.
Changes are against each experiment's matched baseline on its recorded states.}
\label{tab:activation_effects}
\vspace{2pt}
\setlength{\tabcolsep}{5pt}
\begin{tabular}{llccc}
\toprule
Intervention & Donor & $\Delta$ all & $\Delta$ \textsc{Play} & $\Delta$ \textsc{Discard} \\
\midrule
Output transfer (layer 24) & oracle & $+0.091$ & $+0.046$ & $+0.219$ \\
Attention transfer (layer 24) & oracle & $+0.071$ & $+0.033$ & $+0.184$ \\
Restatement transfer (layer 26) & non-oracle & $+0.057$ & $+0.051$ & $+0.069$ \\
\addlinespace
Head scaling (24.29, $\times 1.5$) & none & $+0.021$ & $+0.047$ & $-0.053$ \\
Attention boost & none & $+0.053$ & $+0.100$ & $-0.102$ \\
Averaged directions & none & $+0.027$ & $+0.054$ & $-0.060$ \\
Contrast injection & none & $+0.004$ & $+0.002$ & $+0.009$ \\
Linear map & none & $+0.007$ & $+0.003$ & $+0.018$ \\
\bottomrule
\end{tabular}
\vspace{-10pt}
\end{table}


The best single-layer transfer in a sweep of all 36 layers improves intent accuracy for both action classes: layer 24 for oracle action donors and layer 26 for hint-restatement donors.
Transferring oracle attention weights alone also helps.
Fixed transformations do not reproduce this benefit.
Contrast injection and linear maps change accuracy by less than 0.01, while head scaling, the attention boost, and averaged directions raise \textsc{Play} accuracy but lower \textsc{Discard} accuracy, shifting action preference.
Query replacement and a minimal-context self-donor likewise fail.
Linear readouts of attention-block outputs at the answer position do not recover the convention-consistent intent decision from the \texttt{loop} run at any layer (0.58--0.63 versus a majority rate of 0.767), whereas they do from the oracle action run from layer 7 (0.966).
The hint type and predicted convention are decodable from \texttt{loop} at 0.92 and 0.86; combining these readouts outside the model reaches 0.81.
The tested readouts thus recover the hint type and predicted convention more reliably than the resulting intent decision at these attention-block outputs.
Donor definitions, intervention settings, and detailed results appear in Appendix~\ref{app:activation_interventions}.

\subsection{Learning the Application Step}
\label{sec:lora}

Since the missing combination is not recovered by the tested fixed interventions, we ask whether it can be learned.
We fine-tune a LoRA adapter on Qwen3-8B using 4,000 receiving decisions from the probe training boards, where NOTES contains a randomly sampled rule statement and the label is the target slot and intent decision computed from that rule and the current hint.
Neither ground-truth conventions nor probe predictions are used for training.
The adapter applies stated rules almost perfectly: under \texttt{told}, intent doing rises from 0.599 to 0.998 and game score from 6.16 to 8.43, near the 8.45 ceiling.
It also follows incorrect rule statements, indicating that it has learned the rule-application step itself.
With the original probe, however, \texttt{loop} falls to 4.98 as probe accuracy degrades on the adapter's hidden states.
Retraining the probe raises the score to 6.47, and omitting the poorly decoded target statement raises it further to 6.72.
The remaining limitation therefore shifts from applying the convention to recovering it. The details are in Appendix~\ref{app:lora}.
\section{Conclusion}
\label{sec:discussion}

We studied how LLMs infer and use a partner's communication convention when responding to hints.
Linear probes decoded the convention from receiver representations, but decisions did not follow it.
Stating it as a rule had limited effect; linking it to the current decision helped more, though probe errors bounded the gains.
Convention readout, behavioral adaptation, and game score therefore require separate evaluation.
Using inferred partner information in decisions remains an open challenge for cooperation with unfamiliar partners.

\bibliography{references}
\bibliographystyle{plainnat}

\clearpage

\appendix

\section{Evaluated Models}
\label{app:models}

We evaluate eight open-weight instruction-tuned LLMs, and we list their architecture in~\cref{tab:models-arch} and their source in~\cref{tab:models-source} so that a reader can load the same weights.
Every model is loaded from its Hugging Face repository with the Transformers library (version 5.15.1) on PyTorch 2.6.0 with eager attention, and without a beginning-of-sequence token prepended to the prompt.
For Llama-3.1-8B we load the NousResearch mirror, whose weights are identical to the gated Meta release.
Gemma-3-12B is distributed as a vision-language model, and we use only its language decoder, so the vision tower never receives input.

The probe input at one token position concatenates the residual stream of every layer, counting the embedding output as layer 0, so its dimension is the number of layers plus one times the hidden size.

\begin{table}[t]
\centering\small
\caption{Architecture of the evaluated LLMs.
Heads give the query and key-value head counts, parameters count every weight in the released checkpoint, and the probe input is the length of the concatenated all-layer state at one token position.}
\label{tab:models-arch}
\vspace{2pt}
\setlength{\tabcolsep}{4pt}
\begin{tabular}{lcccccc}
\toprule
Model & Architecture & Layers & Hidden size & Heads (Q/KV) & Parameters & Probe input \\
\midrule
Qwen3-8B & Qwen3 & 36 & 4096 & 32/8 & 8.19B & 151{,}552 \\
Qwen3-32B & Qwen3 & 64 & 5120 & 64/8 & 32.76B & 332{,}800 \\
R1-7B & Qwen2 & 28 & 3584 & 28/4 & 7.62B & 103{,}936 \\
Llama-3.1-8B & Llama & 32 & 4096 & 32/8 & 8.03B & 135{,}168 \\
Tulu-3-8B & Llama & 32 & 4096 & 32/8 & 8.03B & 135{,}168 \\
Hermes-3-8B & Llama & 32 & 4096 & 32/8 & 8.03B & 135{,}168 \\
Gemma-3-12B & Gemma3 & 48 & 3840 & 16/8 & 12.19B$^{\dagger}$ & 188{,}160 \\
Phi-4 & Phi3 & 40 & 5120 & 40/10 & 14.66B & 209{,}920 \\
\bottomrule
\end{tabular}

\vspace{2pt}
\raggedright\footnotesize $^{\dagger}$11.77B in the language decoder, the rest in the unused vision tower.
\end{table}

\begin{table}[t]
\centering\small
\caption{Source of the evaluated LLMs.
Each checkpoint is pinned to a Hugging Face repository and the first eight characters of its commit hash.}
\label{tab:models-source}
\vspace{2pt}
\begin{tabular}{lll}
\toprule
Model & Hugging Face repository & Revision \\
\midrule
Qwen3-8B & \texttt{Qwen/Qwen3-8B} & \texttt{b968826d} \\
Qwen3-32B & \texttt{Qwen/Qwen3-32B} & \texttt{9216db57} \\
R1-7B & \texttt{deepseek-ai/DeepSeek-R1-Distill-Qwen-7B} & \texttt{916b56a4} \\
Llama-3.1-8B & \texttt{NousResearch/Meta-Llama-3.1-8B-Instruct} & \texttt{d10aef79} \\
Tulu-3-8B & \texttt{allenai/Llama-3.1-Tulu-3-8B} & \texttt{66694379} \\
Hermes-3-8B & \texttt{NousResearch/Hermes-3-Llama-3.1-8B} & \texttt{896ea440} \\
Gemma-3-12B & \texttt{google/gemma-3-12b-it} & \texttt{96b6f1ec} \\
Phi-4 & \texttt{microsoft/phi-4} & \texttt{2db69c1c} \\
\bottomrule
\end{tabular}
\end{table}

\section{Environment and Measurement Constants}
\label{app:constants}

We fixed every constant in this appendix before measurement, and we report them here so that a reader can reproduce any number in the main text.

Episodes are organized into disjoint boards by the seed formula
\begin{equation}
\text{seed} = 2{,}300{,}000 + c \times 100{,}000 + i,
\label{eq:seed}
\end{equation}
where $c$ indexes the four convention configurations and $i$ indexes the board.
The four configurations enumerate the convention family, leftmost or rightmost target crossed with color or rank as the play signal, so a board index appears once per configuration and every reported cell averages over all four.
The agent's own convention in the controlled games comes from a separate function of the same seed, using bits that are independent of the partner's convention, so the agent's own convention carries no information about the convention it has to infer.

The board ranges in~\cref{tab:constants} never overlap.
We train probes on the training boards, report every readout number on the test boards, choose every readout constant and every game-level threshold on the constant-selection boards, and run every reported game comparison on the main stage boards.
The matched-state analyses of~\cref{sec:convention_response,sec:activation_effects} use recorded receiving states from the main stage boards and from the constant-selection boards, as listed in Appendix~\ref{app:activation_interventions}.

The environment is a three-color, five-rank variant, so a game scores from 0 to 15 points.
We pair every score difference on the seed, which removes the variance that comes from the deal, and we compute bootstrap 95\% confidence intervals by resampling games rather than decisions, because decisions within one game are not independent.

For the code-based ceiling of~\cref{tab:doing}, the receiving LLM is replaced by a deterministic receiver that applies the sender's assigned target and intent conventions to each received hint.
The scripted sending policy and all other game mechanics remain unchanged.
The resulting score is therefore a ceiling for receiving under the fixed sending policy, not the maximum game score of 15.

\begin{table}[t]
\centering\small
\caption{Design constants, fixed before measurement.
The seed formula of~\cref{eq:seed} makes the board ranges disjoint.}
\label{tab:constants}
\vspace{2pt}
\begin{tabular}{ll}
\toprule
Probe training boards & indices 0--59 per configuration \\
Probe test boards & indices 500--535 per configuration \\
Main stage boards & indices 3000--3059 per configuration (240 games) \\
Constant-selection boards & indices 2000--2059 per configuration \\
Convention family & $2 \times 2$ (target $\times$ intent), drawn per episode \\
Probe & linear logistic on all layers concatenated at one token position \\
Score scale & 0--15, paired seeds, bootstrap 95\% intervals \\
\bottomrule
\end{tabular}
\end{table}

\section{Prompt and Probe Details}
\label{app:format}

\subsection{Prompt Format and NOTES}

A decision prompt has four fixed parts, the rules and the agent's own sending convention, the interaction log, a notes block, and the current board, followed by the question and the two options.
Every line is serialized to a width fixed by the game configuration rather than by its content, so a log line can be replaced by a blank line of the same width.
Fixed-width serialization does not by itself guarantee token alignment, so for every comparison that depends on matched positions we tokenize both prompts and verify that they agree token for token outside the designed difference.
The notes block is present in every condition and holds blank filler when a condition writes nothing into it, which is what lets us compare a written sentence against its absence at matched positions.

The readout input is the prompt body that the action prompt extends, so the probe reads a state that the model computes on the way to its answer, although not the answer position itself.
A variant that appends a reading question to the prompt raises target accuracy to 0.753 on scripted interactions and 0.617 in games, but it is read from a state the model never answers from, so we do not use it.

Every condition other than \texttt{raw} writes four lines into the notes block, and the remaining notes lines keep their blank filler.
The rule-statement conditions, \texttt{loop} and \texttt{told}, write two lines per axis.
\begin{tcolorbox}[breakable, colback=gray!4, colframe=gray!55, colbacktitle=gray!18, coltitle=black,
  fonttitle=\small\bfseries, fontupper=\footnotesize, left=4pt, right=4pt, title={Rule statement (\texttt{loop} and \texttt{told})}]
\begin{Verbatim}[breaklines, breaksymbolleft={}, breakindent=0pt]
  - When P1's hint marks more than one of your cards,
  - the one P1 means is the leftmost of the marked ones
  - When P1 gives you a rank hint, P1 wants you to play
  - the card it means; a colour hint means discard it
\end{Verbatim}
\end{tcolorbox}
The statement names neither the current hint nor a slot, so the model has to apply it to the hint it has just received.
The action-translation conditions, \texttt{loop + verb} and \texttt{told + verb}, write these four lines in place of the rule statement.
\begin{tcolorbox}[breakable, colback=gray!4, colframe=gray!55, colbacktitle=gray!18, coltitle=black,
  fonttitle=\small\bfseries, fontupper=\footnotesize, left=4pt, right=4pt, title={Action translation (\texttt{loop + verb} and \texttt{told + verb})}]
\begin{Verbatim}[breaklines, breaksymbolleft={}, breakindent=0pt]
  - Of the cards this hint marked, the one P1 means
  - is the leftmost of them, which is slot 1
  - For this hint, what P1 wants you to do with
  - that card is to play it
\end{Verbatim}
\end{tcolorbox}
Here external code applies the convention to the current hint and fills in the end, the slot number, and the action.
In both forms, the values come from the probe predictions in \texttt{loop} and \texttt{loop + verb} and from the sender's assigned convention in \texttt{told} and \texttt{told + verb}, and the left end, the rank-play mapping, and the action above are replaced by their alternatives when the convention points the other way.
The boxes quote the notes lines as they appear in the prompt without their fixed-width padding, keeping the British spelling of the prompt text.

\subsection{Answer Interface}

We read answers through two interfaces.
The single-token interface takes the answer as the logit difference between the two option tokens at a fixed position of one forward pass, without sampling and without parsing, and the option order is counterbalanced within every pair.
The generation interface wraps the prompt in the model's chat template, lets the model write freely, and parses a final line into a choice.

Every number in the main text comes from the single-token interface.
Probing and patching take the state at a matched token position as part of their definition, and under free generation the model-written prefix differs across samples, so the matched position that these measurements require does not exist there.
We therefore use the generation interface only as a check, packaging the same decisions for free generation and repeating the readout, and we report numbers from the two interfaces side by side rather than pooling them, because the chat template and the generated prefix change what the model conditions on.

\subsection{Event-Aggregated Readout}

The event-aggregated readout selects the received-side lines of the history, which are the hints the partner gave to this receiver and the actions this receiver took.
Lines recording hints this receiver sent, and the partner's reactions to them, are excluded, because those lines follow this receiver's own convention rather than the partner's.
Selection uses only observable records, never the sender's true convention or unrevealed cards.
At a given turn, we sum the probe logits over the selected lines up to that turn and predict the convention by the sign of the sum.
An individual line need not identify the convention, and because the line representations share preceding context, we do not treat their predictions as independent evidence.
When no line is selected, as at the first turn, the event-aggregated readout makes no prediction.

\subsection{Probe Training}

Each probe standardizes every feature with the mean and standard deviation of the training set and fits an $L_2$-regularized logistic regression, with inverse regularization strength $C=0.01$, by L-BFGS for at most 300 iterations.

\section{Prompt Examples}
\label{app:prompt-examples}

We reproduce two prompts verbatim, both written from the receiver's perspective, in which the receiver is always P0 and its partner P1.
The first box is the readout input at turn 12 of a scripted interaction on probe training board 0 of the first configuration, where the partner's convention is the rightmost target with rank hints meaning play, and the probe reads its final token.
The second box is a receiving decision at turn 18 of a raw game played by Qwen3-8B on board 3000 of the same configuration, where the partner's convention is the leftmost target with rank hints meaning play.
It omits the rule text shared with the first box and shows the target question and the action question, each asked in its own forward pass on the same prompt body.
The notes block reads \texttt{none} in both prompts because neither setting writes into it, and the trailing dots are fixed-width padding.
The model picks slot 1, the leftmost marked card as the partner's convention implies, and then plays it on a color hint, as its own convention implies rather than the partner's.
The remaining conditions use the same prompt body and differ only in the NOTES lines of Appendix~\ref{app:format}.

\begin{tcolorbox}[breakable, colback=gray!4, colframe=gray!55, colbacktitle=gray!18, coltitle=black,
  fonttitle=\small\bfseries, fontupper=\footnotesize, left=4pt, right=4pt, title={Scripted interaction, readout input at turn 12}]
\begin{Verbatim}[breaklines, breaksymbolleft={}, breakindent=0pt]
This game is played by P0 and P1. You are P0.
No player can see their own cards; every player sees every other player's hand.
Your hand has 5 slots, numbered 0 to 4 from left to right.
A card is playable if its rank is exactly one above the current stack of its colour.
Playing a playable card advances that colour's stack. Playing an unplayable card costs a life.
When you play or discard the card in slot k, that card leaves your hand, every card to its
right moves one slot to the left, and a freshly drawn card takes the rightmost slot. So a
slot number names a position in your hand, not a fixed card.
A hint names one colour or one rank and marks every card of the receiver's hand that matches.
Cards that are not marked are known not to match.
In the log, marks [0,1,-,-,4] means slots 0, 1 and 4 are marked and slots 2 and 3 are not.
Near the end of the game, after the deck runs out, hands shrink; `x` marks a slot that no
longer exists, which is different from `-`, a slot whose card simply does not match the hint.
Where card ids appear, each card keeps its id for as long as it stays in the hand, so ids let
you follow a card across slot changes; ids reveal nothing about a card's colour or rank.
LOG LEGEND. Each line is one action. T07 is turn number 7; turns go around the players in
order. Colours are R (red), Y (yellow), G (green). A card is written as its colour
letter and rank, so Y1 is the yellow 1. Verbs: hint = give a hint; play = play the
card in a slot; disc = discard the card in a slot. After play, -> OK means the card
was playable and advanced its stack, -> BAD means it was not and cost a life. A
discard line shows -> -- because a discard has no success or failure. Trailing dots
are padding with no meaning.
Each partner is assigned rules of its own that are consistent, never stated in the
game log, and never change
within a game. Different partners may use different rules.
When a partner gives you a hint, which of the marked cards it means follows that
partner's rule, and whether it asks you to play or to discard that card depends on
the hint being a colour or a rank hint, per that partner's rule.
When you give a partner a hint, it is meant to trust it and act on it on its next turn, whenever
that move is legal: which marked card it takes you to mean, and whether it plays or
discards that card depending on your hint being a colour or a rank hint, also follow
that partner's rules.
Infer each partner's rules from that partner's own hints and responses so far.
A hint has two sides: what its sender means by it follows the sender's rules; what
its receiver takes it to mean under its own rules follows those rules. The two can differ.
Your own moves in the log are meant to follow YOUR RULES below. Under them, your
hints use your play kind aimed at your end when that card is playable, or the
other kind at your end when that card is already on the stacks, as a discard
signal; when a partner hints you, you take the end YOUR RULES name and play it
on your play kind or discard it otherwise, whenever legal; your other discards
pick one of your unmarked slots at random (any slot if all are marked), and with
no legal discard you play a random slot. YOUR RULES describe that assigned
habit, not a guarantee that every logged move obeyed it; when you choose now,
weigh how each partner encodes and reads hints by its own rules.
Each hand line in BOARD lists one other player's cards left to right; `xx` is a slot
that no longer exists.

YOUR RULES
  - When you hint, you mean the leftmost marked card . . . .
  - You signal play with colour hints; rank means discard . . . . .

HISTORY
T01 P1 hint RANK  1 to P0 marks [0,-,2,-,-] . . . . . .
T02 P0 disc slot 0 -> --  card G1 . . . . . . . . . . . . . .
T03 P1 hint RANK  1 to P0 marks [-,1,-,-,4] . . . . . .
T04 P0 disc slot 1 -> --  card G1 . . . . . . . . . . . . . .
T05 P1 hint RANK  1 to P0 marks [-,-,-,3,-] . . . . . . .
T06 P0 disc slot 3 -> --  card G1 . . . . . . . . . . . . . .
T07 P1 play slot 4 -> BAD card G5 . . . . . . . . . . . . . . .
T08 P0 play slot 0 -> BAD card R5 . . . . . . . . . . . . . . .
T09 P1 play slot 0 -> BAD card R3 . . . . . . . . . . . . . . .
T10 P0 play slot 3 -> BAD card R2 . . . . . . . . . . . . . . .
T11 P1 hint RANK  1 to P0 marks [-,-,-,-,4] . . . . . . .

NOTES
  - none . . . . . . . . . . . . . . . . . . .
  - none . . . . . . . . . . . . . . . . . . .
  - none . . . . . . . . . . . . . . . . . . .
  - none . . . . . . . . . . . . . . . . . . .
  - none . . . . . . . . . . . . . . . . . . .
  - none . . . . . . . . . . . . . . . . . . .
  - none . . . . . . . . . . . . . . . . . . .
  - none . . . . . . . . . . . . . . . . . . .

BOARD
stacks R0 Y0 G0 | hints left 7 | lives 11 | deck 13
P1 hand R4 G3 Y5 R3 Y1
\end{Verbatim}
\end{tcolorbox}

\begin{tcolorbox}[breakable, colback=gray!4, colframe=gray!55, colbacktitle=gray!18, coltitle=black,
  fonttitle=\small\bfseries, fontupper=\footnotesize, left=4pt, right=4pt, title={Raw game with Qwen3-8B, receiving decision at turn 18}]
\begin{Verbatim}[breaklines, breaksymbolleft={}, breakindent=0pt]
[Rules and log legend, identical to the first box]

YOUR RULES below. Under them, your
hints use your play kind aimed at your end when that card is playable, or the
other kind at your end when that card is already on the stacks, as a discard
signal; when a partner hints you, you take the end YOUR RULES name and play it
on your play kind or discard it otherwise, whenever legal; your other discards
pick one of your unmarked slots at random (any slot if all are marked), and with
no legal discard you play a random slot. YOUR RULES describe that assigned
habit, not a guarantee that every logged move obeyed it; when you choose now,
weigh how each partner encodes and reads hints by its own rules.
Each hand line in BOARD lists one other player's cards left to right; `xx` is a slot
that no longer exists.

YOUR RULES
  - When you hint, you mean the leftmost marked card . . . .
  - You signal play with colour hints; rank means discard . . . . .

HISTORY
T01 P1 hint RANK  1 to P0 marks [-,-,2,3,-] . . . . . .
T02 P0 play slot 2 -> OK  card G1 . . . . . . . . . . . . . .
T03 P1 hint RANK  1 to P0 marks [-,-,2,-,-] . . . . . . .
T04 P0 play slot 2 -> OK  card R1 . . . . . . . . . . . . . .
T05 P1 disc slot 1 -> --  card R4 . . . . . . . . . . . . . .
T06 P0 hint COLOR R to P1 marks [0,-,-,-,-] . . . . . . . . . .
T07 P1 play slot 0 -> OK  card R2 . . . . . . . . . . . . . .
T08 P0 disc slot 0 -> --  card Y2 . . . . . . . . . . . . . .
T09 P1 hint RANK  2 to P0 marks [-,-,-,3,-] . . . . . . .
T10 P0 play slot 3 -> OK  card G2 . . . . . . . . . . . . . .
T11 P1 disc slot 2 -> --  card Y1 . . . . . . . . . . . . . .
T12 P0 hint RANK  1 to P1 marks [-,1,-,-,4] . . . . . .
T13 P1 disc slot 1 -> --  card G1 . . . . . . . . . . . . . .
T14 P0 disc slot 1 -> --  card Y3 . . . . . . . . . . . . . .
T15 P1 hint RANK  3 to P0 marks [0,-,-,-,-] . . . . . . .
T16 P0 play slot 0 -> OK  card G3 . . . . . . . . . . . . . .
T17 P1 hint COLOR R to P0 marks [-,1,2,-,-] . . . . . . . . .

NOTES
  - none . . . . . . . . . . . . . . . . . . .
  - none . . . . . . . . . . . . . . . . . . .
  - none . . . . . . . . . . . . . . . . . . .
  - none . . . . . . . . . . . . . . . . . . .
  - none . . . . . . . . . . . . . . . . . . .
  - none . . . . . . . . . . . . . . . . . . .
  - none . . . . . . . . . . . . . . . . . . .
  - none . . . . . . . . . . . . . . . . . . .

BOARD
stacks R2 Y0 G3 | hints left 6 | lives 15 | deck 10
P1 hand Y3 G5 R5 Y1 G3
\end{Verbatim}
\par\vspace{2pt}\noindent\rule{\linewidth}{0.4pt}\par\vspace{2pt}
\textit{Target question (first forward pass)}
\begin{Verbatim}[breaklines, breaksymbolleft={}, breakindent=0pt]
P1 just marked slots 1,2. Which one of them do you act on?
OPTIONS
A) SLOT 1
B) SLOT 2

Answer with one letter only.
ANSWER:
\end{Verbatim}
\textit{Model answer A (slot 1)}
\par\vspace{2pt}\noindent\rule{\linewidth}{0.4pt}\par\vspace{2pt}
\textit{Action question (second forward pass, same body)}
\begin{Verbatim}[breaklines, breaksymbolleft={}, breakindent=0pt]
P1 just marked slots 1,2. Will you play or discard slot 1?
OPTIONS
A) DISCARD SLOT 1
B) PLAY SLOT 1

Answer with one letter only.
ANSWER:
\end{Verbatim}
\textit{Model answer B (play slot 1)}
\end{tcolorbox}

\section{Turn Curves}
\label{app:turncurve}

The two readouts do not produce a prediction under the same conditions.
The last-token readout produces one at every turn, since the final token of the decision prompt always exists.
The event-aggregated readout produces one only when at least one received-side line is present, which never holds at the first turn and holds for half of the perspectives at the second turn, because the perspective that moves first has not yet received anything.
Availability reaches 0.94 by the eighth turn, as~\cref{tab:availability} shows.
\Cref{fig:turncurve} therefore restricts both curves to the turns at which both readouts produce a prediction, since reporting the event-aggregated readout over its own turns would compare it against an easier sample.

\begin{table}[!htb]
\centering\small
\caption{Availability of the event-aggregated readout by turn.
Each entry is the fraction of perspectives at that turn for which at least one received-side line is present.}
\label{tab:availability}
\vspace{2pt}
\begin{tabular}{lcccccc}
\toprule
Turn & 1 & 2 & 3 & 4 & 6 & 8 \\
\midrule
Scripted interactions & 0.000 & 0.500 & 0.538 & 0.632 & 0.833 & 0.896 \\
Controlled games & 0.000 & 0.500 & 0.535 & 0.652 & 0.810 & 0.940 \\
\bottomrule
\end{tabular}
\end{table}

The final-turn columns of~\cref{tab:knowing} are not the rightmost points of~\cref{fig:turncurve}.
Each perspective contributes at its own last turn, and those last turns are spread across the horizontal axis.
In controlled games, 150 of 480 perspectives end at turn 33 and 128 end at turn 34, while only 4 reach turn 36, so the rightmost point of the figure averages over those 4 perspectives.

We compute every curve on matched observations and average over observations rather than over perspectives.
The released tables give, for each turn, the axis, the readout, the sample restriction, the accuracy under both averaging rules, the number of correct predictions, the number of observations, and the numbers of distinct perspectives and games, together with the bootstrap interval.

\section{Per-Model Doing and Game Score under Every Condition}
\label{app:model-conditions}

\Cref{tab:model-conditions-a,tab:model-conditions-b} give, for every LLM, the game score and doing under the five LLM receiving conditions of~\cref{tab:doing}, excluding the code-based ceiling, and under the two threshold variants of Appendix~\ref{app:activation_interventions}, on the main stage boards with 240 games per condition, and the \texttt{raw} rows are the values plotted in~\cref{fig:deployment}.
The threshold is the per-model median of the target readout logit magnitude on the constant-selection boards.

\begin{table}[t]
\centering\small
\caption{Doing and game score under every condition, Qwen family and others.
Chance is 0.500 for doing and the score is on the 0--15 scale, higher is better.}
\label{tab:model-conditions-a}
\vspace{2pt}
\setlength{\tabcolsep}{5pt}
\begin{tabular}{llccc}
\toprule
Model & Condition & Score & Doing (target) & Doing (intent) \\
\midrule
\multirow{7}{*}{Qwen3-8B} & \texttt{raw} & 5.63 & 0.515 & 0.591 \\
 & \texttt{loop} & 5.80 & 0.531 & 0.612 \\
 & \texttt{loop} + threshold & 5.95 & 0.545 & 0.629 \\
 & \texttt{loop + verb} & 6.14 & 0.510 & 0.760 \\
 & \texttt{loop + verb} + threshold & 6.28 & 0.552 & 0.753 \\
 & \texttt{told} & 6.16 & 0.821 & 0.599 \\
 & \texttt{told + verb} & 7.78 & 0.962 & 0.867 \\
\midrule
\multirow{7}{*}{Qwen3-32B} & \texttt{raw} & 2.09 & 0.506 & 0.293 \\
 & \texttt{loop} & 2.52 & 0.519 & 0.349 \\
 & \texttt{loop} + threshold & 3.31 & 0.521 & 0.427 \\
 & \texttt{loop + verb} & 6.31 & 0.517 & 0.824 \\
 & \texttt{loop + verb} + threshold & 5.73 & 0.514 & 0.730 \\
 & \texttt{told} & 2.94 & 0.939 & 0.402 \\
 & \texttt{told + verb} & 8.37 & 0.959 & 0.985 \\
\midrule
\multirow{7}{*}{R1-7B} & \texttt{raw} & 4.72 & 0.479 & 0.511 \\
 & \texttt{loop} & 4.98 & 0.500 & 0.533 \\
 & \texttt{loop} + threshold & 4.83 & 0.482 & 0.521 \\
 & \texttt{loop + verb} & 4.92 & 0.480 & 0.562 \\
 & \texttt{loop + verb} + threshold & 4.75 & 0.482 & 0.541 \\
 & \texttt{told} & 5.02 & 0.530 & 0.535 \\
 & \texttt{told + verb} & 5.35 & 0.533 & 0.591 \\
\midrule
\multirow{7}{*}{Gemma-3-12B} & \texttt{raw} & 6.51 & 0.547 & 0.682 \\
 & \texttt{loop} & 6.54 & 0.561 & 0.681 \\
 & \texttt{loop} + threshold & 6.51 & 0.555 & 0.680 \\
 & \texttt{loop + verb} & 6.56 & 0.464 & 0.764 \\
 & \texttt{loop + verb} + threshold & 6.67 & 0.504 & 0.746 \\
 & \texttt{told} & 6.49 & 0.610 & 0.667 \\
 & \texttt{told + verb} & 7.87 & 0.965 & 0.811 \\
\midrule
\multirow{7}{*}{Phi-4} & \texttt{raw} & 3.96 & 0.489 & 0.460 \\
 & \texttt{loop} & 4.18 & 0.481 & 0.478 \\
 & \texttt{loop} + threshold & 4.28 & 0.483 & 0.483 \\
 & \texttt{loop + verb} & 4.08 & 0.447 & 0.548 \\
 & \texttt{loop + verb} + threshold & 4.06 & 0.470 & 0.537 \\
 & \texttt{told} & 4.29 & 0.511 & 0.482 \\
 & \texttt{told + verb} & 4.78 & 0.714 & 0.620 \\
\bottomrule
\end{tabular}
\end{table}

\begin{table}[t]
\centering\small
\caption{Doing and game score under every condition, Llama family.
Chance is 0.500 for doing and the score is on the 0--15 scale, higher is better.}
\label{tab:model-conditions-b}
\vspace{2pt}
\setlength{\tabcolsep}{5pt}
\begin{tabular}{llccc}
\toprule
Model & Condition & Score & Doing (target) & Doing (intent) \\
\midrule
\multirow{7}{*}{Llama-3.1-8B} & \texttt{raw} & 6.62 & 0.500 & 0.698 \\
 & \texttt{loop} & 6.65 & 0.498 & 0.703 \\
 & \texttt{loop} + threshold & 6.68 & 0.498 & 0.709 \\
 & \texttt{loop + verb} & 6.62 & 0.489 & 0.754 \\
 & \texttt{loop + verb} + threshold & 6.46 & 0.500 & 0.743 \\
 & \texttt{told} & 6.59 & 0.508 & 0.699 \\
 & \texttt{told + verb} & 7.25 & 0.588 & 0.818 \\
\midrule
\multirow{7}{*}{Tulu-3-8B} & \texttt{raw} & 6.51 & 0.491 & 0.683 \\
 & \texttt{loop} & 6.65 & 0.494 & 0.700 \\
 & \texttt{loop} + threshold & 6.57 & 0.487 & 0.701 \\
 & \texttt{loop + verb} & 6.54 & 0.460 & 0.776 \\
 & \texttt{loop + verb} + threshold & 6.58 & 0.480 & 0.767 \\
 & \texttt{told} & 6.60 & 0.505 & 0.693 \\
 & \texttt{told + verb} & 7.56 & 0.708 & 0.846 \\
\midrule
\multirow{7}{*}{Hermes-3-8B} & \texttt{raw} & 4.39 & 0.495 & 0.496 \\
 & \texttt{loop} & 4.46 & 0.507 & 0.506 \\
 & \texttt{loop} + threshold & 4.39 & 0.498 & 0.497 \\
 & \texttt{loop + verb} & 5.48 & 0.451 & 0.692 \\
 & \texttt{loop + verb} + threshold & 5.19 & 0.476 & 0.639 \\
 & \texttt{told} & 4.49 & 0.534 & 0.501 \\
 & \texttt{told + verb} & 6.65 & 0.859 & 0.771 \\
\bottomrule
\end{tabular}
\end{table}

\section{Statement Sensitivity Analyses}
\label{app:convention_response}

\subsection{Matched-State Statement Reversals}

We evaluate 800 recorded receiving states, including 348 color-hint states and 452 rank-hint states.
Within each comparison, we keep the history, the slot selected by the unassisted receiver, the question, and the answer options fixed.
The rule-statement conditions differ only in whether color or rank hints mean \textsc{Play}.
The action-recommendation conditions instead state that the current hint means \textsc{Play} or \textsc{Discard}.
The paired rule statements have equal token lengths and differ at two token positions.
The paired action recommendations also have equal lengths and differ at one position.
Option A is fixed to \textsc{Play} and option B to \textsc{Discard} in this diagnostic.

Let $z(x)=\ell_A(x)-\ell_B(x)$ denote the next-token logit margin between the two answer labels.
We define the convention-aligned change as
\begin{equation}
    \Delta_{\mathrm{conv}}
    =
    s_t
    \left[
        z(x_{\mathrm{color\text{-}play}})
        -
        z(x_{\mathrm{rank\text{-}play}})
    \right],
    \qquad
    s_t =
    \begin{cases}
        +1, & \text{for a color hint},\\
        -1, & \text{for a rank hint}.
    \end{cases}
\end{equation}
A positive value indicates movement in the direction implied by the convention change.
The expected direction depends on the current hint type, not on which convention is the ground truth.
For the action recommendations, we define
\begin{equation}
    \Delta_{\mathrm{act}}
    =
    z(x_{\mathrm{play}})
    -
    z(x_{\mathrm{discard}}).
\end{equation}
Choice changes indicate whether the higher-logit answer differs between the paired statements.

The mean convention-aligned change is 0.066, with a game-level bootstrap 95\% confidence interval of $[0.053, 0.079]$.
Its mean is 0.157 for color hints and $-0.005$ for rank hints.
The median absolute logit change is 0.125 for rule statements and 1.250 for action recommendations.
Action-recommendation changes follow the recommended action in all 800 states.
Because this diagnostic fixes the answer order, these results characterize statement sensitivity under that interface rather than performance averaged over option orders.

\subsection{Numerical Precision}

We check the sensitivity of small convention effects to numerical precision on 120 states.
Recomputing the logits in FP32 changes the mean aligned convention effect from 0.1062 to 0.1025 and the action-recommendation effect from 1.4990 to 1.4968.
However, the fraction of exactly zero convention changes falls from 27.5\% to zero.
Among the 33 states with zero changes in the lower-precision measurements, the FP32 median absolute change is 0.0426, and 63.6\% of changes follow the convention-implied direction.
These measurements support the difference in effect magnitude but do not support interpreting lower-precision zeros as complete insensitivity.

\section{Activation Interventions and Additional Diagnostics}
\label{app:activation_interventions}

\subsection{Scope and Donor Conditions}

These experiments analyze the intent decision of Qwen3-8B.
Within each matched-state comparison, the recipient and donor use the same receiving state, selected slot, action question, and option order.
The base model remains frozen throughout.
Ground-truth conventions are used for oracle donors and retrospective evaluation, but not to construct prediction-based interventions.

The recipient run, denoted $L$, receives the probe-predicted convention as a rule statement.
The oracle action donor, denoted $A$, instead receives the target slot and intent decision obtained by applying the true convention to the current hint.
The restatement donor retains prediction-based convention information and restates information about the current hint.
It receives neither the ground-truth convention nor an externally computed action.

The oracle-transfer analysis uses 899 states from boards 3000--3019 for which the donor and recipient inputs have matching token lengths, a condition that excludes no decision there.
The development analyses use 926 states from boards 2000--2019.
The fitted linear-map result is measured on a 452-state test split within the development data.
These state sets are distinct, so their effect sizes should not be interpreted as a ranking under a common evaluation distribution.
Layer 24 and the two heads of the contrast injection were selected on the same main stage states that report them, so those values are selection values, whereas the 206-head set, the linear maps, and the restatement transfer were selected on the constant-selection boards.
Paired differences carry 95\% bootstrap intervals over boards where the record has them.
We did not test borrowing the attention of other heads, attention temperature, fixed additive corrections before the softmax, answer prefill, or transferring only the queries or only the keys of a donor run.

\subsection{Donor Activation Transfers}

We replace the recipient's concatenated attention-head output at the answer position with the corresponding donor output, before the output projection.
We transfer each of the 36 layers in turn and report the layer with the largest gain, layer 24 for the oracle donor and layer 26 for the restatement donor, and the layer order is the same on the main stage and the constant-selection boards.
Replacing all head outputs at layer 24 with those from the oracle action donor raises intent accuracy from 0.603 to 0.694.
The paired change is 0.091, with a bootstrap 95\% confidence interval of $[0.072, 0.112]$.
The intervention corrects 91 decisions and introduces nine errors.
The oracle donor itself has an accuracy of 0.822 on these states.
The same layer-24 transfer on the 926 development states gives 0.094, and transferring the 206 heads of seven layers selected by transfer gain reaches 0.814, the level of the oracle donor itself at 0.806.
Transferring the oracle rule donor, which writes the true convention as a rule statement, gives 0.010 at layer 24 with interval $[0.001, 0.019]$ and nothing elsewhere.

We also separate the donor's attention weights from its value vectors.
Using the oracle donor's attention weights with the recipient's value vectors improves accuracy by 0.071.
Restoring the recipient's attention weights at the NOTES positions, with the corresponding normalization, leaves an improvement of 0.069.
Thus, increased attention to NOTES is not necessary for most of the improvement observed in this particular transfer comparison.
These effects do not establish that attention weights are free of action information, since the donor has already received the oracle action recommendation.

The non-oracle restatement donor provides a further positive comparison.
Transferring all attention-head outputs at layer 26 improves intent accuracy by 0.057 on the 926 development states, with a bootstrap 95\% confidence interval over boards of $[0.035, 0.080]$, and the full restatement run itself improves it by 0.078 with interval $[0.047, 0.108]$.
It corrects 69 decisions and introduces 16 errors, with improvements on both action classes.
This intervention requires a separate restatement execution for each decision.
Its result is therefore distinct from applying a fixed vector estimated across states.

\subsection{Recipient-Side Transformations}

\paragraph{Head-output scaling.}
We multiply selected recipient head outputs by a scalar $g$.
For a head output $o_h$, the intervention is
\begin{equation}
    o'_h = g o_h.
\end{equation}
Experiments include scaling head 24.29 and applying a shared scale to a set of 206 heads selected through donor-transfer analyses.
The shared-scale experiment evaluates
$g \in \{0.5, 0.7, 1.3, 1.5, 2.0\}$.
Its best setting corrects only two additional decisions out of 926.
Individual-head scaling can raise aggregate accuracy by favoring \textsc{Play}, while reducing accuracy on \textsc{Discard} states.

\paragraph{Action preferences under head scaling.}
We evaluate output scaling of head 24.29 on 2,717 receiving decisions under the convention-loop condition.
The reference action is \textsc{Play} in 2,044 decisions and \textsc{Discard} in 673 decisions.
Unlike the statement-reversal diagnostic, this experiment follows the game's randomized option-order procedure and uses the same assigned order across scaling conditions.
A scale of 1.0 leaves the head output unchanged.

\begin{table}[t]
\centering
\small
\caption{Head-output scaling on 2,717 receiving decisions, head 24.29 of Qwen3-8B under the convention loop.}
\label{tab:action_preference}
\vspace{2pt}
\setlength{\tabcolsep}{6pt}
\begin{tabular}{lcc}
\toprule
Measurement & Scale 1.0 & Scale 1.5 \\
\midrule
Overall intent accuracy & 0.584 & 0.605 \\
Accuracy on \textsc{Play} states & 0.694 & 0.741 \\
Accuracy on \textsc{Discard} states & 0.248 & 0.195 \\
Fraction of \textsc{Play} choices & 0.709 & 0.757 \\
\bottomrule
\end{tabular}
\end{table}

Increasing the scale benefits decisions requiring \textsc{Play} while harming those requiring \textsc{Discard}.
Since approximately 75.2\% of decisions require \textsc{Play}, the majority-class improvement outweighs the minority-class loss in aggregate accuracy.
This increase therefore does not by itself establish improved application of the partner's convention.

\paragraph{Convention-contrast injection.}
For each state, we obtain head outputs under the predicted convention and its reversal, denoted $o_h^{p}$ and $o_h^{o}$.
The two inputs share the same answer order and differ only in the rule statement.
We intervene with
\begin{equation}
    o'_h
    =
    o_h^{p}
    +
    \alpha_h
    \left(o_h^{p}-o_h^{o}\right).
\end{equation}
For head 24.29 with $\alpha_h=1$, intent accuracy changes by 0.004 across 2,717 decisions, with a 95\% confidence interval of $[-0.002, 0.010]$.
The intervention corrects 41 decisions and introduces 30 errors.

For head 26.25 with $\alpha_h=-1$, the intervention replaces the predicted-convention output with the opposite-convention output.
It corrects and harms 39 decisions each.
Applying the two interventions together corrects and harms 38 decisions each.
Equal aggregate accuracy therefore does not imply that individual decisions remain unchanged.
The direction has little to carry, because reversing the rule statement barely moves the answer logit at any layer, as the layer sweep below shows, so the difference between the two head outputs is close to noise.

\paragraph{Fitted linear maps.}
We fit linear transformations of the recipient's 128-dimensional head outputs to approximate the corresponding oracle donor outputs.
We examine maps with and without an intercept, low-rank variants of ranks 3 and 8, and selected sets of heads.
The base LLM weights remain unchanged, although the linear transformations are fitted using development data.

Some maps explain approximately 20--50\% of the donor--recipient difference variance on their test split.
However, the best tested intervention yields only three additional correct decisions out of 452, corresponding to an accuracy change of approximately 0.007.
The reported uncertainty interval includes zero.
Reconstructing part of the activation difference therefore does not establish recovery of its behavioral effect.

\paragraph{Averaged restatement directions.}
We compute the answer-position residual difference between a run with hint restatement and the corresponding convention-loop run.
The differences are averaged separately by hint type and applied to other states.
We evaluate individual layers between 6 and 30 and a condition that intervenes at five layers together.

The averaged directions do not reproduce the restatement donor's improvements on both action classes.
Earlier-layer interventions primarily favor \textsc{Play}, whereas later-layer interventions primarily favor \textsc{Discard}.
Similar directional changes occur when the vectors are applied to the unassisted receiver.
The average intervention therefore does not preserve the behavioral effect of transferring the state-specific restatement execution.
The combination that the restatement produces is not stored at the answer position as a fixed direction but arises when the restated hint type is read together with the rule statement and the history through attention, so its average over states keeps only the action-preference component.

\subsection{Additional Mechanistic Diagnostics}

\paragraph{Layer sweep.}
On 160 states we replace the head outputs of one layer in the run that states rank hints mean play with those of the run that states color hints mean play.
Reversing the rule statement moves $z$ by 0.105 on average and changes the choice in 13\% of states, and no layer carries more than 0.045 of that shift, with layer 26 the largest; injecting large random vectors at the same positions moves $z$ by 1.4 to 2.8, so the hook is not at fault.

\paragraph{Attention boost.}
We add a constant to the attention logits from the answer position toward the hint-type token of the current hint line, or toward the whole line, before the softmax.
The best setting, a boost of 8 at layers 20--30 toward the type token, gives $+0.053$ $[0.031, 0.075]$ on the 926 development states, but it raises the rate of choosing \textsc{Play} from 0.722 to 0.823, gains 0.100 on \textsc{Play} states and loses 0.102 on \textsc{Discard} states, and moves the unassisted receiver in the same direction, while boosting the whole line reduces accuracy.
Making the answer position read the hint-type token more is not the same as combining the meaning of that token with the stated rule, so the boost changes how much is read and not what is concluded.

\paragraph{Query replacement.}
Replacing the query at the answer position with the query of the rule statement or of the hint-type token, at layer 24 or at seven layers, on 45 states of one board, lowers accuracy from 0.689 to between 0.40 and 0.64, indistinguishable from a control that uses the query of the board line.

\paragraph{Self-donor with minimal context.}
A donor run of the same model that keeps only the current hint line in the history and the same notes block reaches 0.572 against 0.617 for the recipient, and transferring its head outputs reduces accuracy, by $-0.016$ $[-0.040, 0.007]$ at layer 24 and $-0.049$ $[-0.081, -0.015]$ over seven layers.

\subsection{Why the Tested Fixed Transformations Do Not Reproduce the Transfer}

For the heads that contribute most to the oracle action transfer, the difference between donor and recipient outputs is 10 to 25\% of the output size, 79 to 89\% of its variance lies in three directions, and along those directions the donor output equals the recipient output with slope 0.96 to 1.05 and no offset.
The difference is therefore a state-specific residual rather than a rescaling, and scalar scaling explains none of it under cross-validation.
A $128\times128$ map per head explains 11 to 49\% of that variance on held-out boards, and 57 to 59 of 224 heads exceed 0.2, yet the mapped run makes the same choices as the recipient.

Ridge readouts of the attention-block outputs, trained with four-fold cross-validation over boards on the 926 development states against a majority rate of 0.767, decode the convention-consistent intent decision from the recipient run at 0.58 to 0.63 at every layer, and from the oracle action run at 0.966 from layer 7 and 1.000 from layer 19.
The intent decision obtained by combining the predicted convention with the hint type outside the model, which agrees with the true one at 0.865, is likewise not decodable from the recipient run, at 0.58 to 0.65.
The two premises are decodable from the recipient run, the hint type at 0.92 and the convention at 0.86 at layer 24, combining the two readouts by their agreement gives 0.81, and a direct linear readout of the intent decision gives 0.60, so the premises reach the answer position and their combination does not.
An RBF-kernel readout of the intent decision gives 0.76, at the majority rate.
These readouts cover attention-block outputs at the answer position and are linear, so they leave open information held in the residual stream or the MLP outputs, at earlier positions, or in nonlinear form.

\subsection{Restating the Hint Type}

Outside the game prompt, with only the rule statement, the current hint in words, and the question, 40 prompts covering every convention, hint type, slot, and option order, the model answers at 0.975 with the game wording of the rule.
Adding the pieces of the game prompt one at a time, the drop to 0.550 comes when the hint is written in the format of a history line, and restating the hint type in the question restores 1.000 with every other piece of the game prompt present.
In games on the main stage boards, restating the type of the current hint in the action question raises doing on the intent axis of \texttt{loop} from 0.612 to 0.682 and its score from 5.80 to 5.97 $[-0.04, +0.38]$, but lowers the score of \texttt{raw} from 5.63 to 5.22 $[-0.63, -0.18]$ by moving its rate of choosing \textsc{Play} from 0.730 to 0.634.
The gain of the rule statement given the restatement is $+0.75$ $[+0.52, +0.99]$, while the restatement alone is a bias, so we do not use it as a method.

\subsection{Confidence Threshold on the Target Axis}

On the constant-selection boards the loop follows its target sentence almost regardless of whether it is right, at target doing 0.800 when the readout is right and 0.127 when it is wrong.
We therefore tested omitting the target sentence when the magnitude of the readout logit is below its median on those boards, 2.93 for Qwen3-8B, with a per-LLM median for the other LLMs.
On the main stage boards this raises the score of \texttt{loop} from 5.80 to 5.95 $[-0.05, +0.35]$, which is $+0.33$ $[+0.13, +0.53]$ over \texttt{raw}, and that of \texttt{loop + verb} from 6.14 to 6.28 $[-0.07, +0.35]$, while on Gemma-3-12B, which does not follow the sentence, it changes nothing.
Deriving the threshold from the same quantities on the constant-selection boards gives no threshold, because target doing without any target sentence is 0.414 there, and applying the threshold on those boards lowers target doing from 0.524 to 0.482.
We therefore report the threshold variants with the per-model conditions of Appendix~\ref{app:model-conditions} and do not treat the threshold as a method.

\section{Learning the Application Step}
\label{app:lora}

\subsection{Training Data and Objective}

The adapter is trained on receiving decisions recorded while the base Qwen3-8B plays \texttt{raw} self-play on the probe training boards, 240 games over the four configurations.
At each decision we write one of the four rule statements, leftmost or rightmost target crossed with color or rank meaning play, drawn at random into the NOTES field in the \texttt{loop} wording, and the label is the target slot and intent decision that code derives from that statement and the current hint, so the statement itself is the supervision and neither the assigned convention nor a probe prediction is used.
Prompts in the \texttt{raw} format, without a statement, are labeled with the base model's own answer so that unassisted behavior is preserved.
Of 20,380 recorded lines we use 4,000, 2,000 action and 1,000 slot questions with a statement and 500 of each without.

\subsection{Adapter Configuration}

The adapter has rank 16 and scaling 32 on every attention and MLP projection, 43.6M trainable parameters, and is trained for one epoch of 500 steps with AdamW at learning rate $10^{-4}$ and gradient accumulation 8, with a cross-entropy loss over the two option tokens at the answer position, on one GPU in 1 hour 42 minutes.
The training loss falls from 1.34 to between 0.10 and 0.19.
In games the adapter is merged into the base weights.

\subsection{Probe Retraining}

The base probe loses accuracy on the adapter's hidden states, from 0.865 to 0.774 on intent over the constant-selection states, so we retrain the probes on scripted interactions replayed through the adapter model, which gives 0.921 on intent and 0.650 on target on the probe test boards.
Ground-truth convention labels are used only to train the post-adapter probe and for retrospective scoring, not to train the LoRA adapter.

\subsection{Game Results}

\Cref{tab:lora} gives the game results on the main stage boards.
Because the adapter also changes behavior without a statement, raising \texttt{raw} from 5.63 to 6.21, its conditions are compared with the adapter under \texttt{raw} as well as with the base LLM under \texttt{raw}.
Omitting the target sentence means that the NOTES field holds only the intent statement.

\begin{table}[t]
\centering\footnotesize
\caption{Game score and doing for Qwen3-8B with the application adapter on the main stage boards, 240 games per condition.
No target marks \texttt{loop} with the target sentence omitted, retrained marks the probe trained on the adapter's hidden states, the last two columns are paired differences from the adapter under \texttt{raw} and from the base LLM under \texttt{raw}.}
\label{tab:lora}
\vspace{2pt}
\setlength{\tabcolsep}{3.5pt}
\begin{tabular}{lllcccrr}
\toprule
& & & & \multicolumn{2}{c}{Doing} & & \\
\cmidrule(lr){5-6}
Condition & Model & Probe & Score & Target & Intent & $\Delta$ adapter \texttt{raw} & $\Delta$ base \texttt{raw} \\
\midrule
\texttt{raw} & base & --- & 5.63 & 0.515 & 0.591 & & \\
\texttt{loop} & base & base & 5.80 & 0.531 & 0.612 & & $+0.17$ \\
\texttt{loop}, no target & base & base & 6.06 & 0.503 & 0.645 & & $+0.43$ \\
\texttt{told} & base & --- & 6.16 & 0.821 & 0.599 & & $+0.53$ \\
\texttt{told + verb} & base & --- & 7.78 & 0.962 & 0.867 & & $+2.15$ \\
\addlinespace
\texttt{raw} & adapter & --- & 6.21 & 0.505 & 0.657 & & $+0.58$ \\
\texttt{loop} & adapter & base & 4.98 & 0.573 & 0.651 & $-1.23$ & $-0.65$ \\
\texttt{loop} & adapter & retrained & 6.47 & 0.467 & 0.852 & $+0.25$ & $+0.84$ \\
\texttt{loop}, no target & adapter & base & 4.73 & 0.473 & 0.603 & $-1.48$ & $-0.90$ \\
\texttt{loop}, no target & adapter & retrained & 6.72 & 0.509 & 0.797 & $+0.51$ & $+1.10$ \\
\texttt{told} & adapter & --- & 8.43 & 1.000 & 0.998 & $+2.21$ & $+2.80$ \\
\bottomrule
\end{tabular}
\end{table}

On the constant-selection states the adapter follows the statement it is given at 0.998, so its intent doing under \texttt{loop} equals the probe's accuracy, and it changes sending behavior in ways we have not measured.
The adapter is trained and evaluated on Qwen3-8B only and has not been placed in cross-play.

\section{Additional Discussion and Limitations}
\label{app:discussion}

\noindent\textbf{Access is not application.}
Convention information is decodable (\cref{tab:knowing}), and a true convention stated as a rule changes the intent decision little (\cref{tab:doing}), which points to the step that applies the estimate to the current hint rather than to the estimate itself.
The action-translation conditions perform that step outside the model (\cref{sec:stage}), the donor runs that help carry more explicit decision information (\cref{tab:activation_effects}), and the adapter of~\cref{sec:lora} learns it, so improving the readout alone is unlikely to close the gap.

\noindent\textbf{What counts as adaptation.}
A rule statement produces weak, hint-dependent responses, whereas an action recommendation moves the answer directly (\cref{tab:statement_response}), and an intervention can raise aggregate accuracy by favoring the more frequent action (\cref{sec:convention_response} and Appendix~\ref{app:convention_response}).
Neither a higher accuracy nor a higher game score by itself shows that the receiver became more sensitive to its partner, which is why we report knowing, doing, and game score separately (\cref{sec:setting}) and split intervention effects by states requiring \textsc{Play} and \textsc{Discard} (\cref{tab:activation_effects}).

\noindent\textbf{Location is not content.}
A single layer carries much of the donor's benefit, yet scaling, contrastive, fitted, and averaged transformations at the same layer do not reproduce it (\cref{tab:activation_effects} and Appendix~\ref{app:activation_interventions}).
The tested readouts recover the hint type and the predicted convention more reliably than the resulting intent decision (\cref{sec:activation_effects}), which suggests that an effective intervention needs to supply decision-specific information rather than amplify a fixed direction, although the readouts and transformations we tested do not exclude other forms.

\noindent\textbf{Limitations.}
The environment is a Hanabi-derived two-player game with fixed sender-specific conventions and scripted hint generation, so the results concern controlled receiving decisions rather than autonomous cooperation, and games with more than two players are left to future work.
Within this environment, the probes are trained on scripted interactions, so their transfer to model-generated histories depends on how the available evidence changes, and the donor-transfer gains are measured on recorded states and do not establish gains over complete games.
The mechanistic claims are narrower still, because the activation analyses cover only Qwen3-8B and the readouts and fixed transformations are linear functions of attention-block outputs at the answer position, so information in the residual stream or the MLP outputs, at earlier positions, or in nonlinear form is not excluded.

\end{document}